\documentclass[a4paper]{article}

\usepackage{INTERSPEECH2015}
\usepackage{graphicx}
\usepackage{amssymb,amsmath,bm}
\usepackage{textcomp}
\usepackage{verbatim}
\usepackage[]{algorithm2e}

\usepackage{subfig}
\usepackage{amsmath}
\usepackage{hyperref}
\usepackage{cite}

\usepackage{multirow}

\ninept

\title{Towards Structured Deep Neural Network for Automatic Speech Recognition}

\makeatletter
\def\name#1{\gdef\@name{#1\\}}
\makeatother \name{{\em Yi-Hsiu Liao$^1$, Hung-Yi Lee$^2$, Lin-shan Lee$^3$}}

\address{
  Graduate Institute of Electrical Engineering, National Taiwan University \\
  {\small \tt r03921048@ntu.edu.tw, hungyilee@ntu.edu.tw, lslee@gate.sinica.edu.tw}
}

\begin{document}

\RestyleAlgo{boxruled}

\maketitle
\begin{abstract}

  In this paper we propose the Structured Deep Neural Network (Structured DNN) as a structured and deep learning algorithm, learning to find the best structured object (such as a label sequence) given a structured input (such as a vector sequence) by globally considering the mapping relationships between the structure rather than item by item. 
  When automatic speech recognition is viewed as a special case of such a structured learning problem, where we have the acoustic vector sequence as the input and the phoneme label sequence as the output, it becomes possible to comprehensively learned utterance by utterance as a whole, rather than frame by frame. 
  Structured Support Vector Machine (structured SVM) was proposed to perform ASR with structured learning previously, but limited by the linear nature of SVM. Here we propose structured DNN to use nonlinear transformations in multi-layers as a structured and deep learning algorithm. It was shown to beat structured SVM in preliminary experiments on TIMIT.

\end{abstract}
\noindent{\bf Index Terms}: speech recognition, structured learning, deep neural network, structured deep neural network.

\section{Introduction}

Hidden Markov Models (HMMs\cite{juang1991hidden}) have been the most successful approach for automatic speech recognition for long \cite{rabiner1989tutorial, huang1990hidden}. 
With the maturity of machine learning, great efforts have been made to try to integrate more machine learning concepts into the HMM framework\cite{jiang2006large, collins2002discriminative} because HMMs are generative, while many machine learning approaches can be discriminative in addition\cite{heigold2012discriminative,gales2012structured}. 
Using Deep Neural Networks (DNN)\cite{hinton2006fast, hinton2006reducing} with HMM is a good example\cite{dahl2012context, mohamed2012acoustic, hinton2012deep}.
In general, HMMs consider the phoneme structure by states and the transitions among them, but trained primarily on frame level regardless of being based on DNN\cite{tuske2014acoustic, vinyals2013deep} or Gaussian Mixture Model(or SGMM\cite{povey2010subspace}). 
Under HMM framework\cite{rabiner1993fundamentals}, the hierarchical structure of an utterance is taken care of by the HMM and their states, the lexicon and the language model, which are respectively learned separetely from disjoint sets of knowlage sources. On the other hand, it is well known that there may exist some underlying overall structures for the utterances behind the signals which may be helpful to recognition. If we can learn such structures comprehensively from the signals of the entire utterance globally, the recognition scenario may be different.

On the other hand, structured learning has been substantially investigated in machine learning, which tries to learn the complicated structures exhibited by the data.
Conditional Random Fields (CRF)\cite{lafferty2001conditional, mccallum2003early, gunawardana2005hidden, zweig2009segmental, sung2009hidden, yu2010deep} and structured Support Vector Machine (SVM)\cite{tsochantaridis2005large, zhang2011structured, zhang2013structured} are good example approaches. Recently, structured SVM has been used to perform initial phoneme recognition by learning the relationships between the acoustic vector sequence and the phoneme label sequence of the whole utterance jointly rather than on the frame level or from different sets of knowledge sources\cite{tang2010initial}, utilizing the nice properties of SVM\cite{boser1992training} to classify the structured patterns of the utterance with maximized margin. However, both CRF and structured SVM are linear, therefore limited in analyzing speech signals. Another research\cite{kubo2012integrating} integrates DNN into structured learning but mainly based on Weighted Finite-State Transducers (WFST).

In this paper, we extend the above structured SVM approach to phoneme recognition using a structured DNN including nonlinear units in multi-layers, but similarly learning the global mapping relationships from an acoustic vector sequence to a phoneme label sequence for a whole utterance. Therefore, it is a Structured Deep Neural Network(Structured DNN).

The rest of the paper is organized as follows: Section \ref{sec:arch} is the overall system architecture, Section \ref{sec:struct} introduces structured feature vector, Section \ref{sec:setup} describes details about experiment setup, Section \ref{sec:exp} shows the experiment results, and Section \ref{sec:con} is conclusion and future work.

\section{Proposed Approach -- Structured Deep Neural Network}
\label{sec:arch}
\begin{figure}[t]
  \centering
  \includegraphics[width=\linewidth]{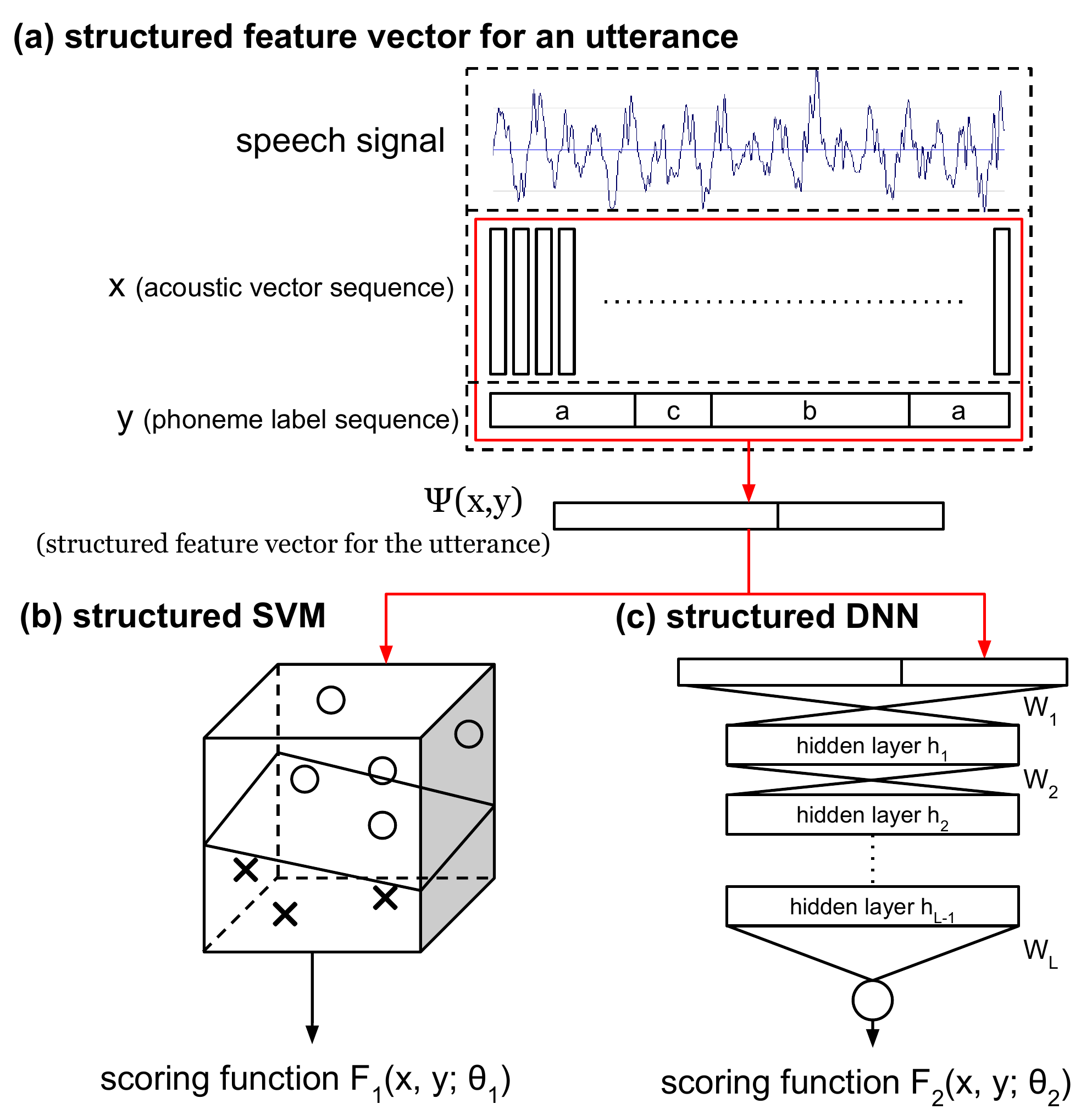}
  \caption{{\it The concept of Structured SVM and Structured Deep Neural Network: (a) the structured feature vector $\Psi(\mathbf{x}, \mathbf{y})$ for an utterance, (b) structured SVM and (c) structured DNN. }} 
  \label{fig:SDNN}
\end{figure}

The whole picture of the concept of the structured DNN for phoneme recognition is in Fig. \ref{fig:SDNN}. Given an utterance with an acoustic vector sequence $\mathbf{x}$ and a corresponding phoneme label sequence $\mathbf{y}$, we can first obtain a structured feature vector $\Psi(\mathbf{x}, \mathbf{y})$ representing $\mathbf{x}$ and $\mathbf{y}$ and the relationships between them as in Fig. \ref{fig:SDNN}(a) (details of $\Psi(\mathbf{x}, \mathbf{y})$ are given in Section \ref{sec:struct}), and then feed it into either an SVM as in Fig. \ref{fig:SDNN}(b) or a DNN as in Fig. \ref{fig:SDNN}(c) to get a score by a scoring function $F_1(\mathbf{x}, \mathbf{y}; \theta_1)$ or $F_2(\mathbf{x}, \mathbf{y}; \theta_2)$, where $\theta_1$ and $\theta_2$ are the parameter sets for the SVM and DNN respectively. Because both $\mathbf{x}$ and $\mathbf{y}$ represent the entire utterance by a structure (sequence) and either SVM or DNN learns to map the pair of $(\mathbf{x}, \mathbf{y})$ to a score on the utterance level globally rather than on the frame level, this is structured learning optimized on the utterance level.

\subsection{Structured Learning Concepts}

In structured learning, both the desired outputs $\mathbf{y}_i$ and the input objects $\mathbf{x}_i$ can be sequences,  trees, lattices, or graphs, rather than simply classes or real numbers.
In the context of supervised learning for phoneme recognition for utterances, we are given a set of training utterances, $(\mathbf{x}_1, \mathbf{y}_1) , ... , (\mathbf{x}_N, \mathbf{y}_N) \in \mathbf{X} \times \mathbf{Y} $ , where $\mathbf{x}_i$ is the acoustic vector sequence of the i-th utterance, $\mathbf{y}_i$ the corresponding reference phoneme label sequence, and we wish to assign correct phoneme label sequences to unknown utterance. 

We first define a function $f(\mathbf{x}; \theta) =\mathbf{y} : \mathbf{X} \rightarrow \mathbf{Y}$ , mapping each acoustic vector sequence $\mathbf{x}$ to a phoneme label sequence $\mathbf{y}$, where $\theta$ is the parameter set be learned. One way to achieve this is to assign every possible phoneme label sequence $\mathbf{y}$ given an acoustic vector sequence $\mathbf{x}$ a score by a scoring function $F(\mathbf{x}, \mathbf{y}; \theta) : \mathbf{X} \times \mathbf{Y} \rightarrow \mathbb{R}$, and take the phoneme label sequence $\mathbf{y}$ giving the highest score as the output of $f(\mathbf{x}; \theta)$,

\begin{equation}
  f(\mathbf{x}; \theta) = \arg\max_{\mathbf{y}\in \mathbf{Y}} F(\mathbf{x}, \mathbf{y}; \theta).
  \label{equ:sl}
\end{equation}

\subsection{Structured SVM}
\label{subsec:ssvm}

Base on the maximized margin concept of SVM, we wish to maximize not only the score of the correct label sequence, but the margin between the score of the correct label sequence and those of the nearest incorrect label sequences, and required the scoring function $F(\mathbf{x}, \mathbf{y}; \theta_1)$ to be linear,

\begin{equation}
  F_1(\mathbf{x}, \mathbf{y}; \theta_1) = \left \langle \theta_1, \Psi(\mathbf{x}, \mathbf{y}) \right \rangle ,
\label{equ:ssvm}
\end{equation}
where $\Psi(\mathbf{x}, \mathbf{y})$ is the structured feature vector mentioned above and shown in Figure \ref{fig:SDNN}, representing the structured relationship between $\mathbf{x}$ and $\mathbf{y}$ , $\theta_1$ is in vector form and $\left \langle \cdot , \cdot \right \rangle $ represents inner product. We can then train the parameter vector $\theta_1$ using training instances $\left \{ (\mathbf{x}_i, \mathbf{y}_i), i = 1, 2,...,\mathbf{L} \right \}$, and then classify the desired label $\mathbf{y}$ for the acoustic vector sequence $\mathbf{x}$ of any unknown testing utterance using the scoring function $F_1(\mathbf{x}, \mathbf{y}; \theta_1)$ with the trained parameter set $\theta_1$. This problem can be solved with the well known SVM\cite{joachims2009cutting}, and is referred to as structured SVM as in Fig. \ref{fig:SDNN}(b).

\subsection{Structured Deep Neural Network (Structured DNN)}
The assumption of the linear scoring function as in (\ref{equ:ssvm}) made structured SVM limited. Instead, the proposed structured DNN uses a series of nonlinear transforms to build the scoring function $F_2(\mathbf{x}, \mathbf{y}; \theta_2)$ with L hidden layers to evaluate a single output value $F_2(\mathbf{x}, \mathbf{y}; \theta_2)$ as in Fig. \ref{fig:SDNN}(c).

\begin{align*}
    \mathbf{h}_1&=\sigma (W_0 \cdot \Psi(\mathbf{x}, \mathbf{y})) &  \\
    \mathbf{h}_l&=\sigma (W_{l-1} \cdot \mathbf{h}_{l-1}),  &2 \leq l \leq L \\
\end{align*}
\vspace{-7mm}
\begin{equation}
  F_2(\mathbf{x}, \mathbf{y}; \theta_2) = \sigma (W_{L} \cdot \mathbf{h}_{L}),
\label{equ:sdnn}
\end{equation}
where $W_i$ is weight matrix (including the bias) of layer i, $\sigma(\cdot)$ a nonlinear transform (sigmoid is used), $h_i$ the output vector of hidden layer i, and the set of all DNN parameters ($W_0$, $W_1$, $W_2$,..., $W_L$) is $\theta_2$. Note that the last weight matrix $W_L$ is a vector, because this DNN gives only a single value as output.

For an utterance with acoustic vector sequence $\mathbf{x} = ( \mathbf{x}^1, \mathbf{x}^2, ..., \mathbf{x}^M)$, where $\mathbf{x}^j$ is the j-th acoustic vector, there can be many possible phoneme label sequences $\mathbf{y} = (y^1, y^2, ..., y^M)$, where $y^j$ is the phoneme label for $\mathbf{x}^j$, and a reference phoneme label sequence $\mathbf{t} = (t^1, t^2,..., t^M)$, where $t^j$ is the true phoneme label for $\mathbf{x}^j$. The label accuracy function $C_x(\mathbf{t}, \mathbf{y})$ for the utterance $\mathbf{x}$ can then be calculated as follows,

\begin{equation}
  C_x(\mathbf{t}, \mathbf{y}) = \frac{1}{M}\sum_{j=1}^{M}{\delta(t^j, y^j)},
  \label{equ:conf}
\end{equation}
where $\delta(t^j, y^j)$ is 1 if $t^j = y^j$, 0 otherwise. $C_x(\mathbf{t}, \mathbf{y})$ in (\ref{equ:conf}) is actually the frame accuracy or one minus the frame error rate. When we are more interested in minimizing the phone error rate, the definition of the label accuracy function $C_x(\mathbf{x}, \mathbf{y})$ in (\ref{equ:conf}) can be modified to reflect that goal. In both cases, the parameter set $\theta_2$ of this DNN is trained by minimizing the following loss function,

\begin{equation}
  L(\theta_2)= -\sum_{\mathbf{x}}{C_x(\mathbf{t}, \mathbf{y})\log{F_2(\mathbf{x}, \mathbf{y}; \theta_2)}},
  \label{equ:loss}
\end{equation}
where $L(\theta_2)$ in (\ref{equ:loss}) is summed over all training utterances, and this objective function is defined in a way similar to the cross entropy popularly used in DNN training.

\subsection{Inference with Structured DNN}

With the structured DNN trained as above, given the acoustic vector sequence $\mathbf{x}$ of an unknown utterance, we need to find the best phoneme label sequence $\mathbf{y}$ for it. For structured SVM in subsection \ref{subsec:ssvm}, due to the linear assumption, the learned model parameter $\theta_1$ contains enough information to execute the Viterbi algorithm to find the best label sequence. This is not true for structured DNN. From (\ref{equ:sl}), in principle we need to search over all possible phoneme label sequences($K^M$ for $K$ phonemes and $M$ acoustic vectors) for the given acoustic vector sequence and pick the one giving the highest score, which is computationally infeasible.

Instead of searching through all possible phoneme label sequences, we can start from a random label sequence, and then change one phoneme label at a time by going through all phoneme labels with all other phoneme labels in the utterance fixed. We iterate the overall phoneme label sequences in this way until the sequence converges. This referred to as "without lattice". We can also decode using WFST first to generate a lattice, and then, choose the phone label sequence from the lattice which give the highest score. Of course in this way, the performance is bounded by the quality of the lattice. This is referred to as "with lattice".

\subsection{Training of Structured DNN}

For each training utterance, again we have $K^M$ possible label sequences. It is also impossible to train over all these label sequences for the training utterances. In structured SVM, there is a large margin training algorithm which finds training examples to produce the maximum margin. For structured DNN here, how to find and choose effective training examples is important. Besides the positive examples (reference phoneme label sequences for the training utterances), in this work negative examples (those other than reference label sequences) are chosen both by random and by inferencing using the current model. The latter is explained below.

The "inferenced label sequences" represent a feedback mechanism. When the current structured DNN model is used to decode a training utterance and obtain a phoneme label sequence, which is far from correct, we add this label sequence to help training data to adjust the model. Wtih the training data generated, standard backpropagation algorithm can be used to update the structured DNN parameters, and additional training sequences can be regenerated in each epoch.
Because inference can be performed with or without lattice, same for training. For training with lattice, we choose N-best paths and N random paths from lattice as the inference label sequences.

\section{Structured Feature Vector \texorpdfstring{$\Psi(\mathbf{x}, \mathbf{y})$} for an utterance}
\label{sec:struct}

\begin{center}
\begin{figure}[t]
  \centering
  \subfloat[a simple example with arbitrary acoustic vector]{{\includegraphics[width=0.17\textwidth]{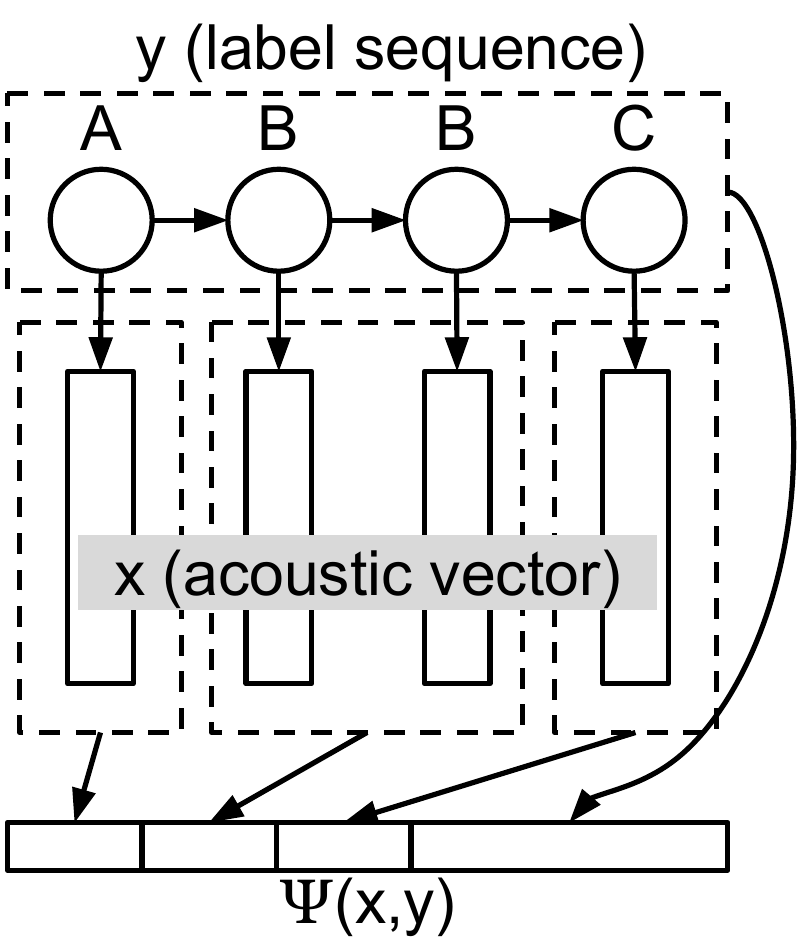} }}%
  \qquad
  \subfloat[a demonstration of how $\Psi(\mathbf{x}, \mathbf{y})$ (not normalized yet) is computed.]{{\includegraphics[width=0.24\textwidth]{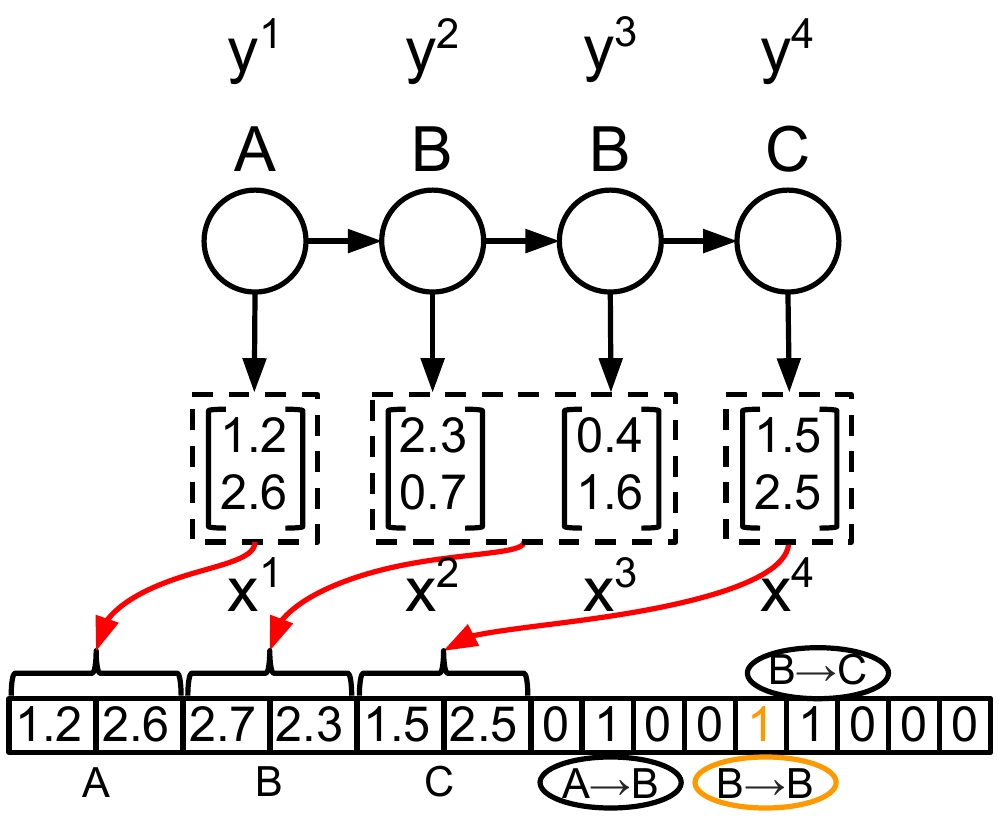} }}%
  \caption{{\it A simplified example of feature sequence $\mathbf{x} = (\mathbf{x}^1, \mathbf{x}^2, \mathbf{x}^3, \mathbf{x}^4 )$ and label sequence $\mathbf{y} = (\mathbf{y}^1, \mathbf{y}^2, \mathbf{y}^3, \mathbf{y}^4 ) = (A, B, B, C) $.}}
  \label{fig:psi_ex}
\end{figure}
\end{center}

Take the MFCC vector or phoneme posteriorgram vectors as the acoustic vectors for an utterance of $M$ frames, $\mathbf{x} = \{\mathbf{x}^j, j = 1, 2, . . . M \}$, and the phoneme label for $\mathbf{x}^j$ is $\mathbf{y}^j$. So the task is to decode $\mathbf{x}$ into the label sequence $\mathbf{y} = \{y^j, j = 1, 2, ... M \}$. 
Since the most successful and well known solution to this problem is with HMM, we try to encode what HMM has been doing into the feature vector $\Psi(\mathbf{x}, \mathbf{y})$ to be used here.
An HMM consists of a series of states, and two most important sets of parameters -- the transition probabilities between states, and the observation probability distribution for each state. 
Such a structure is slightly complicated for the work here, so in the preliminary work we use a simplified HMM with only one state for each phoneme.
With this simplification, these two sets of probabilistic parameters can be estimated for each utterance by adding up all the counts of the transition between labels (or states) and also adding up all the acoustic vectors for each label (phoneme or state), then normalizing the results with the length of the utterance. This is shown in Fig. \ref{fig:psi_ex}(a).

Assume $K$ is the total number of different phonemes, we first define a $K$ dimensional vector $\Lambda(y^j)$ for $y^j$ with its k-th component being 1 and all other components being 0 if $y^j$ is the k-th phoneme. 
Tensor product $\otimes$ is helpful here, which is defined as
\begin{equation}
  \otimes : \mathbb{R}^P \times \mathbb{R}^Q \rightarrow \mathbb{R}^{PQ}, (a\otimes b)_{i+(j-1)P} \equiv a_i \times b_j,
\label{equ:tensor}
\end{equation}
where $\mathbf{a}$ and $\mathbf{b}$ are two ordinary vectors with dimensions $P$ and $Q$ respectively. The right half of (\ref{equ:tensor}) says $\mathbf{a}\otimes \mathbf{b}$ is a vector of dimension $PQ$, whose $\left [ i + (j-1)P \right ]$-th component is the i-th component of $\mathbf{a}$ multiplied by the j-th component of $\mathbf{b}$. With this expression, the feature vector $\Psi(\mathbf{x}, \mathbf{y})$ in Fig. \ref{fig:SDNN}(a) to be used for evaluating the scoring function $F_1(\mathbf{x}, \mathbf{y}; \theta_1)$ in (\ref{equ:ssvm}) or $F_2(\mathbf{x}, \mathbf{y}; \theta_2)$ in (\ref{equ:sdnn}) can then be configured as the concatenation of two vectors,
\begin{equation}
  \label{equ:phi}
  \Psi(\mathbf{x}, \mathbf{y}) = \frac{1}{M}
  \begin{pmatrix}
    \sum_{j=1}^{M}{\mathbf{x}^j\otimes\Lambda(\mathbf{y}^j)}
    \\ 
    \sum_{j=1}^{M-1}{\Lambda(\mathbf{y}^j)\otimes\Lambda(\mathbf{y}^{j+1})}
  \end{pmatrix},
\end{equation}
where $\mathbf{x} = \{\mathbf{x}^1, \mathbf{x}^2,..., \mathbf{x}^M\}$ and $\mathbf{y} = \{y^1, y^2,..., y^M\}$. The upper half of the right hand side of (\ref{equ:phi}) is to accumulate the distribution of all components of $\mathbf{x}^j$ for each phoneme in the acoustic vector sequence $\mathbf{x}$, and then locate them at different sections of components of the feature vector $\Psi(\mathbf{x}, \mathbf{y})$ (corresponding to the observation probability distribution for each state or phoneme label estimated with the utterance). 
The lower half of the right hand side of (\ref{equ:phi}), on the other hand, is to accumulate the transition counts between each pair of labels (phonemes or states) in the label sequence $\mathbf{y}$ (corresponding to state transition probabilities estimated for the utterance). After normalizing with the utterance length $M$ (which is also helpful to give a good range for input of DNN) $\Psi(\mathbf{x}, \mathbf{y})$ is then the concatenation of the two, so it keeps the primary statistical parameters of $\mathbf{x}^j$ for different phonemes $y^j$ for all $\mathbf{x}^j$ in $\mathbf{x}$, and the transitions between states for all $y^j$ in $\mathbf{y}$. With enough training utterances $(\mathbf{x}, \mathbf{y})$ and the corresponding function $\Psi(\mathbf{x}, \mathbf{y})$, we can then learn the scoring function $F_1(\mathbf{x}, \mathbf{y}; \theta)$ or $F_2(\mathbf{x}, \mathbf{y}; \theta_2)$ by training the parameters $\theta_1$ or $\theta_2$.
The vector $\Psi(\mathbf{x}, \mathbf{y})$ in (\ref{equ:phi}) can be easily extended to higher order Markov assumptions (transition to the next state depending on more than one previous states). For example, by replacing the upper half of (\ref{equ:phi}) with $\sum_{n=1}^{N}{\mathbf{x}^n\otimes\Lambda(\mathbf{y}^n)\otimes\Lambda(\mathbf{y}^{n+1})}$ and the lower half of (\ref{equ:phi}) with 
$\sum_{n=1}^{N-1}{\Lambda(\mathbf{y}^n)\otimes\Lambda(\mathbf{y}^{n+1})\otimes\Lambda(\mathbf{y}^{n+2})}$
, we have the second order Markov assumption.

Consider a simplified example for $K = 3$ (only 3 allowed phonemes A, B, C) and an utterance with length $M = 4$ as shown in Fig. \ref{fig:psi_ex}(b). It is then easy to find that the upper half of $\Psi(\mathbf{x}, \mathbf{y})$ is $\sum_{n=1}^{4}{\mathbf{x}^n\otimes\Lambda(\mathbf{y}^n)} = {(1.2, 2.6, 2.7, 2.3, 1.5, 2.5 )}' $, and the lower half of $\Psi(\mathbf{x}, \mathbf{y})$ is $\sum_{n=1}^{3}{\Lambda(\mathbf{y}^n)\otimes\Lambda(\mathbf{y}^{n+1})} = {(0,1,0,0,1,1,0,0,0)}'$. We therefore have $\Psi(\mathbf{x}, \mathbf{y}) =\frac{1}{4} \cdot {(1.2, 2.6, 2.7, 2.3, 1.5, 2.5, 0, 1, 0, 0, 1, 1, 0, 0, 0)}'$ .

\vspace{-10mm}
\section{Experimental Setup}
\label{sec:setup}

Initial experiments were performed with TIMIT. We used the training set without dialect sentences for training and the core testing set (with 24 speakers and no dialect) for testing.
The models were trained with a set of 48 phonemes and tested with a set of 39 phonemes, conformed to CMU/MIT standards\cite{lee1989speaker}.
we used an online library\cite{SSVM_toolkit} for structured SVM, and modified the kaldi\cite{kaldi} code to implement structured DNN. 
\begin{center}
\begin{table*}[!t]
\centering
\begin{tabular}{|l||c|c|c||c|}
\hline
     & \begin{tabular}[c]{@{}c@{}}structured\\ SVM\end{tabular} & \begin{tabular}[c]{@{}c@{}}structured DNN\\ (without lattice)\end{tabular} & \begin{tabular}[c]{@{}c@{}}structured DNN\\ (with lattice)\end{tabular} & \begin{tabular}[c]{@{}c@{}}Karel's\\ recipe\end{tabular} \\ \hline \hline
(a)LDA-MLLT-fMLLR          & 38.62 & 90.50 & 19.98 & 18.90 \\ 
(b)phone post(kaldi)       & 24.32 & 30.62 & \textbf{18.77} & x\\ 
(c)phone post(filter bank) & 30.84 & 40.65 & x & x \\ \hline
\end{tabular}
\caption{Phone Error Rate(\%) evaluate on 39 phonemes with different acoustic vectors, and different algorithms, structured SVM, structured DNN without lattice, structured DNN with lattice (top 2000-best path), and state-of-the-art kaldi results. }
\label{tab:summary}
\end{table*}
\end{center}
\vspace{-5mm}
Our experiment is based on Karel's recipe in kaldi for TIMIT script, which used LDA-MLLT-fMLLR features obtained from auxiliary GMM models, as RBM pre-training, frame cross-entropy training and sMBR. On top of Karel's recipe, we used three sets of acoustic vectors, (a)LDA-MLLT-fMLLR feature (40 dimensions), or input to DNN in Karel's recipe; (b)phoneme posterior probability (48 dimensions) obtained from the 1943 DNN output (state posterior) from Karel's recipe by reducing the dimension to mono-phoneme size of 48 with an additional layer of DNN ($1943 \times 48$); and (c) phoneme posterior probability from filter bank(48 dimensions) obtained by a DNN (4-layer of 512 neurons) with filter bank input of context 4-1-4. They are respectively referred to as acoustic vectors (a)(b)(c) below.  
\section{Experimental Results}
\label{sec:exp}

The results are listed in Table \ref{tab:summary}. The structured DNN without lattice (3 hidden layers 200 neurons per layer) did not perform better than structured SVM, although it did learn some structured patterns, obviously because of the poor quality of the training data (most of them are random), as well as the fact that the inference algorithm of changing only one phoneme label at a time is actually prone to converge at local maximal. On the other hand, training/inferencing with lattice was much better, because it offered many effective training examples by picking up the top N paths from the lattice.
This structured DNN with lattice gave a phone error rate (PER) of 18.77\% which outperformed structured SVM and is actually slightly better than Karel's recipe of kaldi at 18.90\% in Table \ref{tab:summary}. Note that in the experiments here, as explained in Section\ref{sec:struct}, we simply assume a single state for a phoneme in (\ref{equ:phi}), which is certainly over-simplified. 

\begin{figure}[ht]
  \centering
  \includegraphics[width=\linewidth]{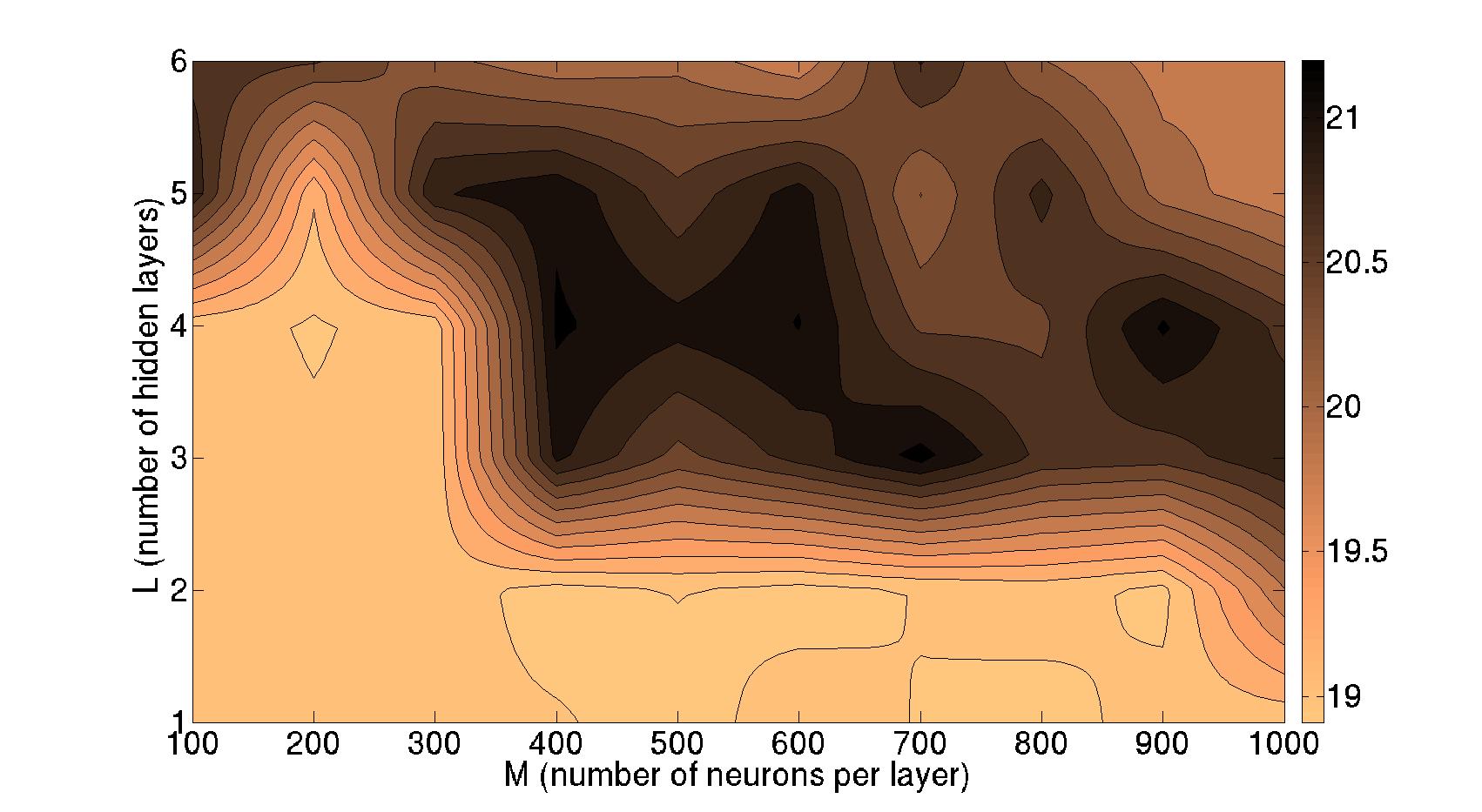}
  \caption{{Phone Error Rate(\%) map for the proposed structured DNN with lattice, N = 500 using acoustic vector (b), for different values of $L$ (number of hidden layers) and $M$ (number of neurons per layer).}}
  \label{fig:dnn}
\end{figure}

\begin{center}
\begin{table}[b]
\centering
\begin{tabular}{|l||c|c|c|}
\hline
           & N = 500  & N = 1000 & N = 2000 \\ \hline \hline
oracle min &  11.25   &  10.67   &  10.10   \\ 
oracle max &  29.52   &  30.70   &  31.82   \\ 
random     &  20.88   &  21.16   &  21.44   \\ 
  SDNN     &  \textbf{18.91}   & \textbf{19.19}   &  \textbf{18.77}   \\ \hline \hline
rand - SDNN&  1.97    &  1.97    &  \textbf{2.07}   \\ \hline
\end{tabular}
\caption{Phone Error Rate(\%) for structured DNN with lattice for different N of the N-best paths from lattice with acoustic vectors (b) in Table \ref{tab:summary}.}
\label{tab:nbest}
\end{table}
\end{center}
\vspace{-7.4mm}
We further used the best acoustic vectors (b) of Table \ref{tab:summary} and change the number N for the N-best path from lattice in training/inferencing, and the results are in Table \ref{tab:nbest}. In the case that we already know the reference phone label sequence, we can choose the path closest to the reference, most different from the reference, or randomly choose one, out of the N-best paths. These gave the oracle min, oracle max, and random, the first three rows in Table \ref{tab:nbest}. For N = 500, 1000 or 2000, the structured DNN(SDNN) row is always better than random in N-best, but with a gap from oracle min. This verified that the structured DNN did learn some structure information. Of course, the oracle min here is a lower bound for structured DNN with lattice proposed here. The last row shows the difference between random and structured DNN, which increases as training data(N) grows, verified that larger data set provides better results.

The next experiment is to analyze the phone error rates (PER) for different choices of the key hyper-parameters for the structured DNN, $L$ number of hidden layers and $M$ number of neurons in each hidden layer.
Figure \ref{fig:dnn} is the result, a visualized PER map for structured DNN with lattice, N = 500 using acoustic vector (b). The horizontal axis is $M$ where $M = 100, 200, ... 1000$, and the vertical axis is $L$ where $L = 1, 2, ... 6$. Therefore, the figure consists of $6\times 10 = 60$ data points. The overall performance is approximately 18\% to 21\%, all outperforming the structured SVM, and more or less comparable to Karel's recipe. 
For this task, better PER seemed to be located at less hidden layers and less neurons. If $M$ is small, we can use deep networks up to 4 hidden layers, whereas $M$ is large, 2 hidden layers is enough. With larger $L$ and larger $M$, DNN overfits training data. The best PER is on $(L, M) = (1, 500)$ which was the case in Table \ref{tab:nbest}.
\section{Conclusion and Future Work}
\label{sec:con}
In this paper, we propose a new structured learning architecture, structured DNN for phoneme recognition which jointly considers the structures of acoustic vector sequences and ph/oneme label sequences globally. Preliminary test results show that the structured DNN out-performed the previously proposed structured SVM and provided a comparable result to state-of-the-art. We will work on multiple states per phone in the future, and explore the possibility of structured DNN.

\newpage
\eightpt
\bibliographystyle{IEEEtran}

\bibliography{mybib}
\end{document}